# A Novel Bayesian Classifier using Copula Functions

Saket Sathe

Electrical Engineering Department

Indian Institute of Technology Bombay

`saket.sathe@gmail.com`

## Abstract

A useful method for representing Bayesian classifiers is through *discriminant functions*. Here, using copula functions, we propose a new model for discriminants. This model provides a rich and generalized class of decision boundaries. These decision boundaries significantly boost the classification accuracy especially for high dimensional feature spaces. We strengthen our analysis through simulation results.

## 1 Introduction

Pattern classification is an important task in several image processing, statistical learning, and data mining applications. The most popular pattern classifiers are Bayesian classifiers. There are many well known methods for representing Bayesian classifiers, but one of the most useful method is by *discriminant functions*. These functions provide inter-class decision surfaces for Bayesian classifiers.

*Discriminant functions* assume several forms depending on the probability density of the feature space. But most attention has been received by discriminant functions that assume multivariate Gaussian distribution [1]. This attention has largely been due to analytical tractability of the multivariate Gaussian distribution. Obviously, a multivariate Gaussian distribution assumes that the marginals are univariate Gaussian distributed. However, this assumption is clearly impertinent in a large number of statistical learning problems.

In this paper, using copula functions, we build a rich and generalized class of discriminant functions. These discriminant functions model the joint dependence structure independent of marginal distributions. We name these discriminant functions as *copula discriminant functions*. We believe, until now copula functions have never been used for pattern recognition. However, they have recently been used in areas like computational finance and computational geology.

Our approach results in a 25% to 45% increase in classification accuracy. Similar to *SVM*s (Support Vector Machines), copula discriminants exhibit exemplary stability at higher dimensions. High dimensional feature spaces typically appear in tasks like text classification, image processing, and natural language processing.

We begin by introducing copula functions. Copula functions (also, copulas) are advanced statistical tool for modeling and estimation of multivariate probability density functions (*pdf*s). In a multivariate distribution estimation setting, copulas enable us to separate the joint dependence structure from the marginal distribution. This elegant separation also allows us to create arbitrary distribution functions with known joint dependence structure, and without imposing restrictions on the marginal distributions. We use this property to derive decision boundaries which are sensitive to marginal distributions.

The rest of the paper is organized as follows: in Section (2) we briefly introduce Bayesian classifiers. Then in Section (3) we introduce copulas and use them to model decision boundaries. Estimation of copula functions is discussed in Section (4). We supplement our approach by simulation results in Section (5). Lastly, we draw conclusions in Section (6).

## 2 Bayesian Decision with Multivariate Normal Discriminant Function

We briefly introduce the Bayesian decision technique for building Bayesian classifiers.

Let $\{\omega_1, \ldots, \omega_c\}$ be a finite set of $c$ states of nature, or classes. Let the feature space be $\mathbf{x} \in \mathbb{R}^d$. The problem of classification is to assign each $\mathbf{x}$ a label $\omega_1, \ldots, \omega_c$. Let $g_i(x)$, $i = 1, \ldots, c$ be discriminant functions, a classifier is said to assign $\mathbf{x}$ a label $\omega_i$ if,

$$g_i(\mathbf{x}) > g_j(\mathbf{x}) \qquad \text{for all } j \neq i. \tag{1}$$

Also, it is worthwhile to note that, for any monotonically increasing function $Q(\cdot)$, $Q(g_i(x))$ will keep the classification unaltered [1]. A well known form of $g_i(x)$ is the *Bayes decision surface*,

$$\begin{aligned} g_i(\mathbf{x}) = P(\omega_i|\mathbf{x}) &= \frac{f(\mathbf{x}|\omega_i)P(\omega_i)}{\sum_{j=1}^c f(\mathbf{x}|\omega_j)P(\omega_j)} \\ g_i(\mathbf{x}) &= \ln f(\mathbf{x}|\omega_i) + \ln P(\omega_i), \end{aligned} \tag{2}$$

where $f(\cdot)$ is the probability density function, $P(\cdot)$ is the probability mass function, and 'ln' denotes natural logarithm. Classifiers obtained from (2) are known as *Bayesian Classifiers*. These classifiers achieve a minimum-error-rate classification [1].

Evaluating (2) for $f(\mathbf{x}|\omega_i) \sim N(\mu_i, \Sigma_i)$ gives us the *normal discriminant function*,

$$g_i(\mathbf{x}) = -\frac{1}{2}(\mathbf{x} - \mu_i)^t \Sigma_i^{-1}(\mathbf{x} - \mu_i) - \frac{d}{2}\ln(2\pi) - \frac{1}{2}\ln|\Sigma_i| + \ln(P(\omega_i)). \tag{3}$$

The assumption that $f(\mathbf{x}|\omega_i) \sim N(\mu_i, \Sigma_i)$ tacitly implies that $x_j \sim N(\mu_{ij}, (\sigma_{jj}^2)_i)$. Thus a normal discriminant cannot accurately classify samples whose joint dependence structure is different from its marginal distribution. Recall, this is mainly due to lack of tractably in such problems.

In the next section we will model discriminant functions using copulas. Copulas allow us to tackle problems mentioned earlier without loosing analytical tractability. Using copulas, joint distributions can be modeled incorporating idiosyncrasies of marginal distributions.

## 3 Copulas for Modeling Multivariate Probability Density Functions

We begin by rigorously defining copulas and their necessary components. Then, using copula functions we propose discriminant functions for Bayes classifiers that are superior and flexible than the multivariate normal density.

**Definition 3.1 (Copula)** *A d-dimensional copula is a function $C : [0,1]^d \mapsto [0,1]$ and having the following properties:*

1. *Let $\mathbf{u} \in [0,1]^d$, thus $\mathbf{u} = \{u_1, \ldots, u_d\}$, where $u_j \in [0,1] \; \forall j \in \{1, \ldots, d\}$, then $C(\mathbf{u})$ is increasing in each component,*

2. *$C(\mathbf{u}) = \mathbf{0}$ if at least one coordinate $u_j = 0$,*

3. *$C(\mathbf{u}) = u_k$ if $u_j = 1 \; \forall j \neq k$,*

4. *For every $\mathbf{a}, \mathbf{b} \in [0,1]^d$ with $\mathbf{a} \leq \mathbf{b}$, and a hypercube $\mathbf{B} = [\mathbf{a}, \mathbf{b}] = [a_1, b_1] \times [a_2, b_2] \times \cdots \times [a_d, b_d]$ whose vertices lie in the domain of $C$, we have volume $V_C(\mathbf{B}) \geq 0$.* [1]

One of the most pivotal theorems in copula theory is Sklar's theorem. It provides a relationship between multivariate distributions and their univariate counterparts.

**Theorem 3.1 (Sklar's theorem)** *Let $F$ be a $d$-dimensional distribution function with margins $F_1, F_2, \ldots, F_d$. Then there exits an $d$-copula $C$ such that for all $x_1, x_2, \ldots, x_d$ in $\mathbb{R}^d$,*

$$F(x_1, x_2, \ldots, x_d) = C(F_1(x_1), F_2(x_2), \ldots, F_d(x_d)). \tag{4}$$

*If $F_1, F_2, \ldots, F_d$ are continuous, then $C$ is unique; otherwise, $C$ is uniquely determined on $\text{Ran}(F_1) \times \text{Ran}(F_2) \cdots \times \text{Ran}(F_d)$.*[2] *Conversely, if $C$ is an $d$-copula and $F_1, F_2, \ldots, F_d$ are distribution functions, then the function $F$ defined by (4) is an $d$-dimensional distribution function with margins $F_1, F_2, \ldots, F_d$.*

---

[1] $V_C(\mathbf{B})$ is defined as follows,

$$V_C(\mathbf{B}) = \sum_{i_1=1}^{2} \cdots \sum_{i_d=1}^{2} (-1)^{i_1 + \cdots + i_d} C(u_{1,i_1}, \ldots, u_{d,i_d})$$

for all $u_{1,1}, \ldots, u_{d,1} \in [0,1]^d$ and $u_{1,2}, \ldots, u_{d,2} \in [0,1]^d$ with $u_{n,1} \leq u_{n,2}$.

[2] $\text{Ran}(F)$ indicates range of the function $F(\cdot)$.

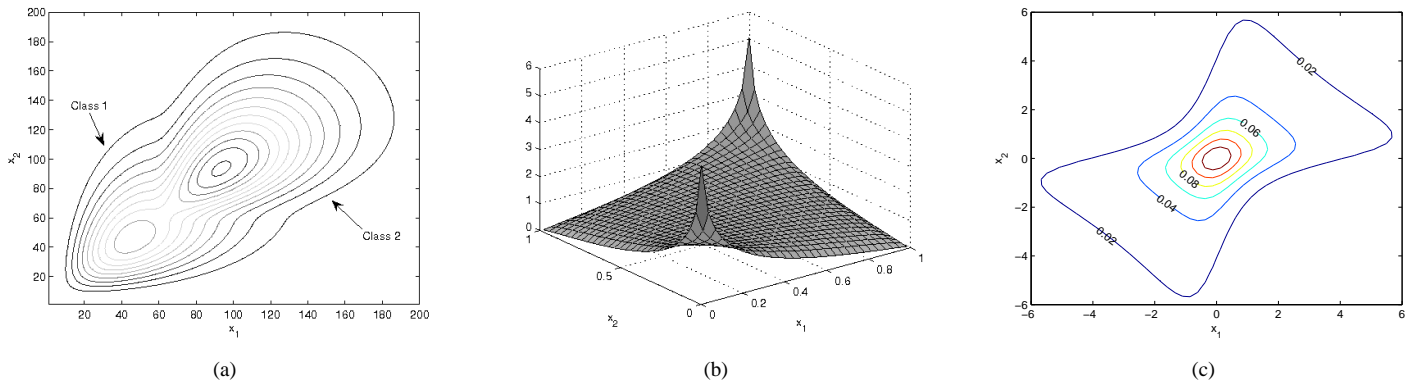

(a) (b) (c)

Figure 1: (a) Contour plot of a two-class two-feature density function where the marginals are gamma distributed but are jointly multivariate normal, (b) Gaussian copula density for $\rho_{12} = 0.4$, (c) Contour plots of marginally Student's t and jointly Gaussian random variables.

*Proof:* Refer [10].

Thus, by Sklar's theorem it is possible to decompose a multivariate distribution function into a copula $C(\cdot)$ and marginal distributions $F_1, \ldots, F_d$. This provides a bottom-up approach for modeling a multivariate distribution function. First, estimate the marginal distributions and then use a copula (Equation 4) to obtain a multivariate distribution.

**Definition 3.2 (Copula Density)** *Copulas do not always have densities. However, most of the copulas we study here possess copula densities. Copula density is given as follows,*

$$c(x_1, \ldots, x_d) = \frac{\partial^d C(F_1(x_1), F_2(x_2), \ldots, F_d(x_d))}{\partial x_1, \cdots, \partial x_d}. \tag{5}$$

*For an absolutely continuous joint distribution F with strictly increasing and continuous marginal dfs $F_1, \ldots, F_d$, we may differentiate $C(x_1, \ldots, x_d)$ to see that the copula density is given by,*

$$\begin{aligned} c(x_1, \ldots, x_d) &= \frac{f(F_1^{-1}(x_1), \ldots, F_d^{-1}(x_d))}{\prod_{k=0}^{d} f_k(F_k^{-1}(x_k))}, \\ c(F_1(x_1), \ldots, F_d(x_d)) &= \frac{f(x_1, \ldots, x_d)}{\prod_{k=1}^{d} f_k(x_k)}, \end{aligned} \tag{6}$$

$$f(x_1, \ldots, x_d) = c(F_1(x_1), \ldots, F_d(x_d)) \prod_{k=1}^{d} f_k(x_k). \tag{7}$$

The knowledge of copula density is particularly useful for estimating parameters of a copula.

Thus a probability density function can be written as a product of copula density and marginal density functions. A copula captures the dependence structure between marginals without imposing restrictions on the marginal distribution. For example, we can easily construct a probability density function wherein $c(\cdot)$ is a Gaussian copula density (refer Figure 1.a), but $F_1, \ldots, F_d$ are non-Gaussian (beta, Poisson, exponential, etc.).

Furthermore, by a processes called *empirical marginal transform* (Section 4), we can transform individual features to obtain their respective *empirical distribution function*. These *empirical distribution functions* can replace the unknown distribution functions. This allows us to further relax assumptions on marginals and leads to a significant increase in classification accuracy.

Now, let us turn our attention to Bayesian classifiers. From (2) we replace $f(\mathbf{x}|\omega_i)$ with $f^i(x_1, \ldots, x_d)$. Thus from (7) we have,

$$f^i(x_1, \ldots, x_d) = c^i(F_1(x_1), \ldots, F_d(x_d)) \prod_{k=1}^{d} f_k(x_k, \theta_k^i), \tag{8}$$

where $\theta_k$ are parameters of the marginals. From (2) we simplify further,

$$\begin{aligned} g_i(\mathbf{x}) &= \ln\{c^i(F_1(x_1), \ldots, F_d(x_d))\} + \ln\{\prod_{k=1}^{d} f_k(x_k, \theta_k^i)\} + \ln P(\omega_i), \\ g_i(\mathbf{x}) &= \ln\{c^i(F_1(x_1), \ldots, F_d(x_d))\} + \sum_{k=1}^{d} \ln\{f_k(x_k, \theta_k^i)\} + \ln P(\omega_i), \end{aligned} \tag{9}$$

if prior probabilities $P(\omega_i)$'s are same for all the classes then,

$$g_i(\mathbf{x}) = \ln\{c^i(F_1(x_1), \ldots, F_d(x_d))\} + \sum_{k=1}^{d} \ln\{f_k(x_k, \theta_k^i)\}. \tag{10}$$

Thus, as shown in (9) and (10) Bayesian classifiers can be derived using copula densities. This gives us flexible and true decision boundaries in terms of discriminant functions. As it can be observed, these discriminant functions are much superior and generalized than their Gaussian counterparts. In Section (5) we shall further strength the above generalization through simulation results.

Copulas exist in various shapes and forms [3, 10]. But, the most useful and popular copulas are the parametric copulas, namely, the Gaussian and the Student's t copula (also, t copula). We define these copulas as follows,

**Definition 3.3 (Gaussian Copula)** *Let $\rho$ be a symmetric, positive definite matrix with $diag(\rho) = 1$ and let $\Phi_\rho$ the standardized multivariate normal distribution with correlation matrix $\rho$. Then the multivariate Gaussian copula is defined as,*

$$C(u_1, \ldots, u_d; \rho) \triangleq \Phi_\rho(\Phi^{-1}(u_1), \ldots, \Phi^{-1}(u_d)), \tag{11}$$

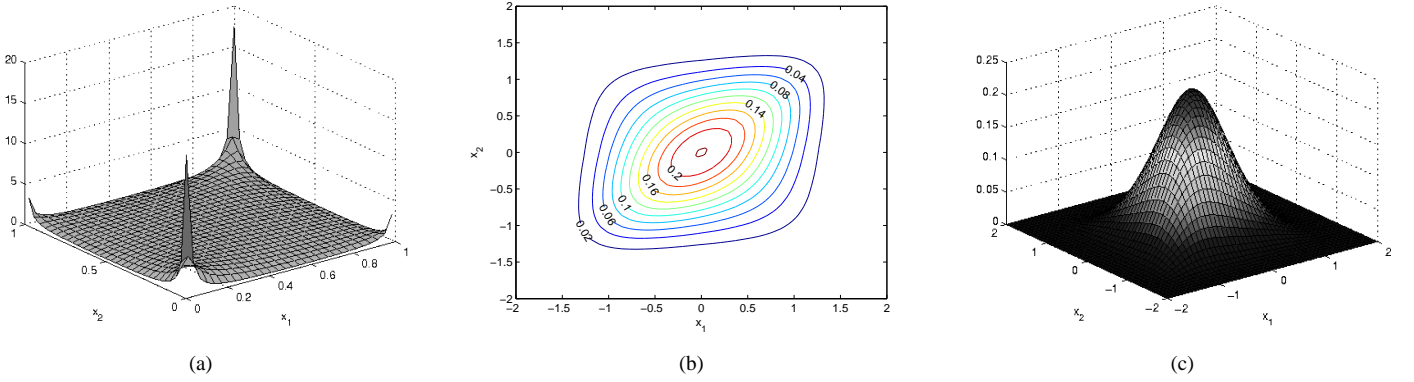

(a) (b) (c)

Figure 2: (a) Student's t copula density for $\rho_{12} = 0.4$, $\nu = 2$. (b) and (c) show contour and surface plots respectively of marginally Gaussian and jointly Student's t random variables

where $\Phi^{-1}(u)$ denotes the inverse of the normal cumulative distribution function. The density function of the Gaussian copula is given as,

$$c(u_1, u_2, \ldots, u_d; \rho) \triangleq \frac{1}{|\rho|^{\frac{1}{2}}} \exp\left[-\frac{1}{2}\zeta'(\rho^{-1} - \mathbb{I})\zeta\right]. \tag{12}$$

where $\zeta = (\Phi^{-1}(u_1), \Phi^{-1}(u_2), \ldots, \Phi^{-1}(u_d))'$. *Figure 1(b) and Figure 1(c) depict various forms of a bivariate Gaussian copula density.*

**Definition 3.4 (Student's t copula)** *Let $\rho$ be a symmetric, positive definite matrix with $diag(\rho) = 1$ and let $T_{\rho,\nu}$ the standardized multivariate Student's t distribution with correlation matrix $\rho$ and $\nu$ degrees of freedom. Then the multivariate Student's t copula is defined as follows,*

$$C(u_1, \ldots, u_d; \rho, \nu) \triangleq T_{\rho,\nu}(t_\nu^{-1}(u_1), \ldots, t_\nu^{-1}(u_d)), \tag{13}$$

*where $t_\nu^{-1}(u)$ denotes the inverse of the Student's t cumulative distribution function. The associated Student's t copula density is obtained by applying equation,*

$$c(u_1, u_2, \ldots, u_d; \rho, \nu) \triangleq |\rho|^{\frac{1}{2}} \frac{\Gamma(\frac{\nu+n}{2})}{\Gamma(\frac{\nu}{2})} \left[\frac{\Gamma(\frac{\nu}{2})}{\Gamma(\frac{\nu+1}{2})}\right]^n \frac{(1 + \frac{\zeta'\rho^{-1}\zeta}{\nu})^{-\frac{\nu+n}{2}}}{\prod_{i=1}^n (1 + \frac{\zeta_i^2}{\nu})^{-\frac{\nu+1}{2}}}, \tag{14}$$

*where $\zeta = (t_\nu^{-1}(u_1), t_\nu^{-1}(u_2), \ldots, t_\nu^{-1}(u_d))'$. Figure (2) depicts various forms of a bivariate Student's t copula density.*

# 4 Estimation of Copulas

There are a few well known methods for copula parameter estimation [2, 6, 9]. Most of these methods involve direct maximization of log-likelihood function of the copula density with respect to the parameters.

## 4.1 Exact Maximum Likelihood (EML) Method

Let $\theta_c \in \Theta$ be the space of parameters. Let $\{\mathbf{x}^{n,i}\}_{n=1}^{N_i} \forall \mathbf{x} \in \mathbb{R}^d$, represent $N_i$ training samples form the $i^{th}$ class, where $i = 1, \ldots, c$. Also, let $\mathcal{L}_n(\theta_c)$ and $\ell_n(\theta_c)$ be the likelihood and the log-likelihood functions of the underlying copula for the $n^{th}$ sample. Then,

$$\ell(\theta_c) = \sum_{n=1}^{n=N_i} \ln c(F_1(x_1^{n,i}), \ldots, F_d(x_d^{n,i}); \theta_c) + \sum_{n=1}^{N_i} \sum_{k=1}^{d} \ln(f_k(x_k^{n,i}, \theta_k^i)). \quad (15)$$

Now, we maximize $\hat{\theta}_c$ with respect to (15) as,

$$\hat{\theta}_c = \arg\max\{\ell(\theta_c) | \theta_c \in \Theta\}.$$

Thus for the Gaussian copula the log-likelihood function takes the form,

$$\ell^{Ga}(\theta_c) = -\frac{N_i}{2} \ln |\rho| - \frac{1}{2} \sum_{n=1}^{N_i} \zeta'_{n,i}(\rho^{-1} - \mathbb{I})\zeta_{n,i}, \quad (16)$$

where $\theta_c = \{\rho | \rho \in \mathbb{R}^{d \times d}\}$ and $\rho$ is a symmetric positive definite matrix. Thus as it can be seen the estimation of a Gaussian copula is simple and the parameter ($\rho$) also have a closed form solution,

$$\hat{\rho}^i = \frac{1}{N_i} \sum_{n=1}^{N_i} \zeta^{n,i} \zeta^{n,i'}. \quad (17)$$

Similarly, we can write $\ell(\theta_c)$ for the Student's t copula as,

$$\begin{aligned}\ell^{St}(\theta_c) &= N_i \ln\left(\frac{\Gamma(\frac{\nu+d}{2})}{\Gamma(\frac{\nu}{2})}\right) - dN_i \ln\left(\frac{\Gamma(\frac{\nu+d}{2})}{\Gamma(\frac{\nu}{2})}\right) \\ &\quad -\frac{N_i}{2} \ln |\rho| - \frac{\nu+d}{2} \sum_{n=1}^{N_i} \ln(1 + \frac{\zeta'_{n,i} \rho^{-1} \zeta_{n,i}}{\nu}) \\ &\quad +\frac{\nu+1}{2} \sum_{n=1}^{N_i} \sum_{k=1}^{d} \ln(1 + \frac{\zeta_{kn,i}^2}{\nu}),\end{aligned} \quad (18)$$

where $\theta_c = \{(\nu, \rho) | \nu \in (2, \infty], \rho \in \mathbb{R}^{d \times d}\}$. But, the maximization of log-likelihood function with respect to the parameters is complicated as it involves simultaneous estimation of the parameters of the dependence structure ($\rho$) and the margins ($\nu$). A more efficient method known as the canonical maximum likelihood method (CML) is proposed by Mashal and Zeevi [9]. We discuss this method in Section (4.2).

| Dataset | Marginal probability density functions | Gaussian copula discriminant (% Accuracy) | Normal discriminant (% Accuracy) | $SVM^{light}$ $C < \frac{1000}{max(\|r\|^2)}$ (% Accuracy) | $SVM^{light}$ $C \in 2^i$ $i \in \{-5, ..., 15\}$ (% Accuracy) |
|---|---|---|---|---|---|
| 1 | $t_2$ | 99.10 | 72.47 | 84.53 | 98.50 |
| 2 | Gam(4,2) | 80.78 | 22.94 | 81.67 | 93.87 |
| 3 | Exp(0.7) | 76.62 | 40.54 | 99.20 | 99.80 |
| 4 | Gam(4.3,1.7), Log-N(0.64,0.22) | 81.86 | 35.06 | 79.93 | 92.73 |
| 5 | Exp(0.6), Gam(4,2) | 78.82 | 34.16 | 69.67 | 89.13 |
| 6 | Log-N(0.7,0.2), Gam(5,3), Exp(0.5) | 79.86 | 39.68 | 62.33 | 90.53 |
| 7 | Exp(0.32), Gam(3.1,4.3), $\chi^2$(3.2) | 77.32 | 40.62 | 61.73 | 93.20 |
| 8 | Log-N(0.53,0.36), Gam(6.2,3.3), Exp(0.44), $\chi^2$(5) | 80.24 | 38.92 | 65.53 | 92.40 |

Table 1: Comparison of Gaussian copula discriminant and normal discriminant for various datasets. All datasets possess a 100-dimensional feature space.

## 4.2 Canonical Maximum Likelihood (CML) Method

The method described above makes assumption on the distribution of the marginals while it performs parameter estimation. However, the CML method does not make any distributional assumption. It uses a transformation known as the *empirical marginal transformation* to transform data to an estimate of its empirical distribution. This estimate of the empirical distribution approximates the unknown marginal distribution $F_k(\cdot)$ as,

$$\hat{F}_k(\cdot) \triangleq \frac{1}{N_i} \sum_{n=1}^{N_i} 1_{\{\mathbf{x}_{kn} \leq \cdot\}}, \tag{19}$$

where $1_{\{\mathbf{x}_{kn} \leq \cdot\}}$ is an indicator function. Thus the empirical marginal transformation transforms the data to uniform variates.

Mashal and Zeevi [9] show the effective application of the CML method for calibrating a Student's t copula. They propose an robust estimator for $\rho$ that use rank correlation, particularly the Kendall's $\tau$. Using this estimator the CML method can be summarized as in Algorithm (4.1):

**Algorithm 4.1 (CML method for Student's t copula)**

1. *Transform the data $\mathbf{x}_n$ to pseudo-samples $\mathbf{u}_n$ using the empirical marginal transformation;*

2. *Estimate the correlation matrix $\rho$ as in [9];*

3. *Perform the unconstrained maximization for $\hat{\nu}$ as,*

$$\hat{\nu} = \arg \max_{\nu \in (2, \infty]} \sum_{n=1}^{N_i} \log c(\hat{u}_1^n, \hat{u}_2^n, \ldots, \hat{u}_d^n; \rho, \nu). \tag{20}$$

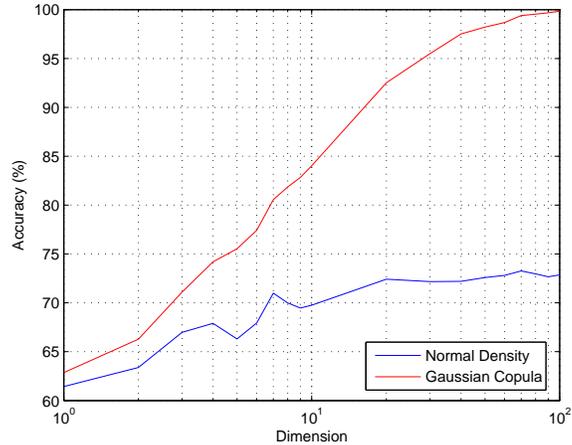

Figure 3: Accuracy of Gaussian copula discriminant

The above method has computational advantages over other methods [9]. This method does not require matrix inversion and therefore has the advantage of being numerically stable in the presence of close-to-singular correlation matrices. The estimator for $\rho$ is $O(n^2)$ for each coefficient. This is better that the iterative procedure for estimation. And, as the maximization is carried out with respect to a single parameter ($\nu$), it is significantly faster and stable.

## 5 Simulation Results

We simplify our problem to a two-category classification problem. Using synthetic datasets we establish the propriety of our approach. Although similar simulation methodology is incorporated for all datasets (see Table 4.1), due to space limitations we present detailed results for only *Dataset-1*. The features of this dataset are individually Student's t distributed but are jointly multivariate normal. This dataset consists of around 4000 samples. We us 70% of the samples for training and 30% samples for prediction.

We start by estimating marginal empirical distribution functions as given in (19). Then we numerically differentiate the empirical distribution functions to get the empirical density functions. It is worthwhile to note that we have not made *any* distributional assumption on marginal densities. Later, using the EML method we estimate parameters of the Gaussian copula density for all classes. Recall, these copula densities lead to copula discriminants, which are then used to predict class labels as seen in Section (2).

For every dimension we evaluate both methods several times. The ensemble average of these evaluations is reported in Figure (3). Clearly, the copula discriminant function is about 25% more accurate at higher dimensions. This inability of the normal discriminant to classify high-dimensional data accurately can be attributed to *Hughes* *phenomenon* [4, 11]. Intuitively, as the dimension increases; accuracy of the normal discriminant should increase, since a new feature would only add more information about the data. As a result, Bayes error is a decreasing function of the dimensionality of data [11]. However, Figure (3) reveals that normal discriminant functions exhibit a degradation of accuracy as the dimension increases. This degradation is caused due to a relative increase in classification error as compared to the decrease in Bayes error [11]. This increase in classification error is due to the fact that more parameters need to be estimated from the same number of training samples. Obviously, if this increase in classification error is greater than the decrease in Bayes error the overall performance degrades [11]. This is known as *Hughes phenomenon*. Lee and Landgrebe [8] argue that at high dimensions the value of second order statistics of a class contain more information as compared to the first order statistics. Admittedly, copula discriminant functions have an unique capability to exploit this higher order information (Figure 3).

In Table (4.1) we summarize simulation results for all datasets. Furthermore, we compare the efficacy of our approach with *SVMs*. We have chosen *SVMs* since they are kernel based non-Bayesian classifiers. Also, they are known to be significantly accurate in a variety of classification tasks.

Each of these dataset bear a 100 dimensional feature space and consist of around 4000 samples. Here, the individual feature distribution can assume various forms (either Gamma, $t_\nu$, Lognormal, exponential, or $\chi^2$) as indicated by the second column in Table (4.1).

For training and classification using *SVMs* we use the $SVM^{light}$ [5] package with an *RBF* kernel. Search for optimal *RBF* kernel parameters is performed using a simple search algorithm given in [7, 12]. We restrict the *SVM* trade-off parameter $C < 1000 \times \left\{max(||r||^2)\right\}^{-1}$ for all feature vectors $r$ where $max(\cdot)$ is the maximum value and $||\cdot||$ indicates Euclidean norm. However, it is also common to use $C = \left\{avg(||r||^2)\right\}^{-1}$ where $avg(\cdot)$ indicates arithmetic average.

We report the accuracy obtained from above methods in Table (4.1). Again, it evinces that the Gaussian copula discriminant function is about 35% to 40% more accurate as compared to the normal discriminant function. The Gaussian copula discriminant also exhibits an accuracy comparable to *SVMs*, although we have observed a better performance for *SVMs* if $\ln_2 C = \{-5, ..., 15\}$ [12]. Accuracy of *SVMs* for various values of $C$ is presented in column five and column six of Table (4.1). Admittedly, copula discriminants have several advantages over *SVMs*:

1. They obviate solving of a quadratic programming problem, as required by *SVMs*. Also, for obtaining the empirical marginal distribution functions we only have to sort the training data *(O(nlogn))*.

2. The memory foot-print of copulas is much smaller than *SVMs*. Only a small subset of points representing the marginal densities need to be stored.

3. Closed form solutions for parameters are available and produce significantly better results. Thus a time consuming search over the parameter space for optimal parameters becomes redundant.

# 6 Conclusion

Copulas provide generalized decision boundaries for Bayesian classifiers. Clearly, this generalization is effective and increases classification accuracy significantly. Most importantly, this elevation in accuracy is not achieved at the expense of analytical tractability. Thus, compared to popular models, copulas provide a superior model for pattern classification. However, the method for choosing a copula that best fits the joint feature distribution is still at the forefront of research. In our subsequent papers we shall explore this issue along with other applications of copulas to statistical learning.

# 7 Acknowledgment

The authors would like to thank Prof. Charles Elkan, Department of Computer Science and Engineering, University of California, San Diego, Prof. Uday Desai, Electrical Engineering Department, IIT Bombay, and Prof. Rajendra Lagu, IIT Bombay for their continual support and valuable suggestions during the course of this work.